\documentclass[final,5p,times]{tempcls}

\usepackage{amsmath,amssymb,amsfonts}
\usepackage{amsopn}
\usepackage{graphicx}
\usepackage{subfigure}
\usepackage{epstopdf}
\usepackage{multirow}
\usepackage{float}
\usepackage{epsfig}
\usepackage{graphicx}
\usepackage{multirow}
\usepackage{amsopn}
\usepackage{array}
\usepackage{arydshln}
\usepackage{amsthm,bm}
\usepackage{times}
\usepackage{multicol}
\usepackage{bm}

\journal{}
\begin{document}
\begin{frontmatter}
\title{Sparse Graph-based Transduction for Image Classification}


\author[a]{Sheng Huang}
\author[a]{Dan Yang\corref{cor1}}
\author[a]{Jia Zhou}
\author[b]{Luwen Huangfu}
\author[c,d]{Xiaohong Zhang}
\address[a]{College of Computer Science at Chongqing University, Chongqing, 400044, P.R.C}
\address[b]{Eller College of Management at University of Arizona, Tucson, AZ, 85712, USA}
\address[c]{School of Software Engineering at Chongqing University, Chongqing, 400044, P.R.C}
\address[d]{Ministry of Education Key Laboratory of Dependable Service Computing in Cyber Physical Society, Chongqing, 400044, P.R.C}
\cortext[cor1]{Corresponding author (Dan Yang): dyang@cqu.edu.cn}


\begin{abstract}
Motivated by the remarkable successes of Graph-based Transduction (GT) and Sparse Representation (SR), we present a novel Classifier named Sparse Graph-based Classifier (SGC) for image classification. In SGC, SR is leveraged to measure the correlation (similarity) of each two samples and a graph is constructed for encoding these correlations. Then the Laplacian eigenmapping is adopted for deriving the graph Laplacian of the graph. Finally, SGC can be obtained by plugging the graph Laplacian into the conventional GT framework. In the image classification procedure, SGC utilizes the correlations, which are encoded in the learned graph Laplacian, to infer the labels of unlabeled images. SGC inherits the merits of both GT and SR. Compared to SR, SGC improves the robustness and the discriminating power of GT. Compared to GT, SGC sufficiently exploits the whole data. Therefore it alleviates the undercomplete dictionary issue suffered by SR. Four popular image databases are employed for evaluation. The results demonstrate that SGC can achieve a promising performance in comparison with the state-of-the-art classifiers, particularly in the small training sample size case and the noisy sample case.
\end{abstract}

\begin{keyword}
Image Classification, Sparse Representation, Graph Learning, Transductive Learning, Semi-supervised Learning
\end{keyword}

\end{frontmatter}

\section{Introduction}
As two popular techniques for classification, Sparse Representation (SR) and Graph-based Transduction (GT) have attracted a lot of attentions in machine learning, computer vision and image processing communities~\cite{src,gsrc,ssc,srdp,ksrc,hyper,zhou}. The idea of SR stems from the compression sensing that most signals have a sparse representation as a linear combination of a reduced subset of signals from the same space~\cite{src,compress}. The basic idea of GT is to utilize the similarities between each two samples to infer the labels of unlabeled samples where such similarities are encoded in a graph or hypergraph~\cite{zhou,st,dhlp,gtam,ssl}. In SR, the signals tend to have a representation biased towards their own class and only the most relevant signals are highlighted~\cite{srdp}. These facts endow SR with the strong discriminating power and the robustness to noise. However, an important prior condition of SR is that it requires the dictionary to be overcomplete. In the lack of training samples case (the undercomplete dictionary case), which actually is very common in the real world applications, the dictionary constructed by training samples is too small to sparsely represent the query sample which will restrict the classification performance of SR. Moreover, another shortcoming of SR is that it cannot utilize the self-similarities of the training data and the self-similarities of the testing data. On the contrary, GT can well alleviate the previous shortcomings suffered by SR, since the graph, which is core of GT and encodes the similarities, are constructed from both training and testing samples.  In other words, all data can be sufficiently exploited. The main problem of the current GT approaches is that they are easily corrupted by noise. This is due to the fact that most of GT approaches generate the graphs (or hypergraphs) by k-nearest-neighbour and $\epsilon$-ball~\cite{sparsegraph}. However, improving the robustness to noise is what is SR good at. Apparently, the advantages SR and GT are complementary. So here comes a question, if there exists a classification approach that can combine SR and GT together and inherit their advantages? Fortunately, this paper will give a positive answer.
\begin{figure*}[!tbp]
\centering
\subfigure[Raw Samples and their rank scores]{
\centering
\includegraphics[scale=0.45]{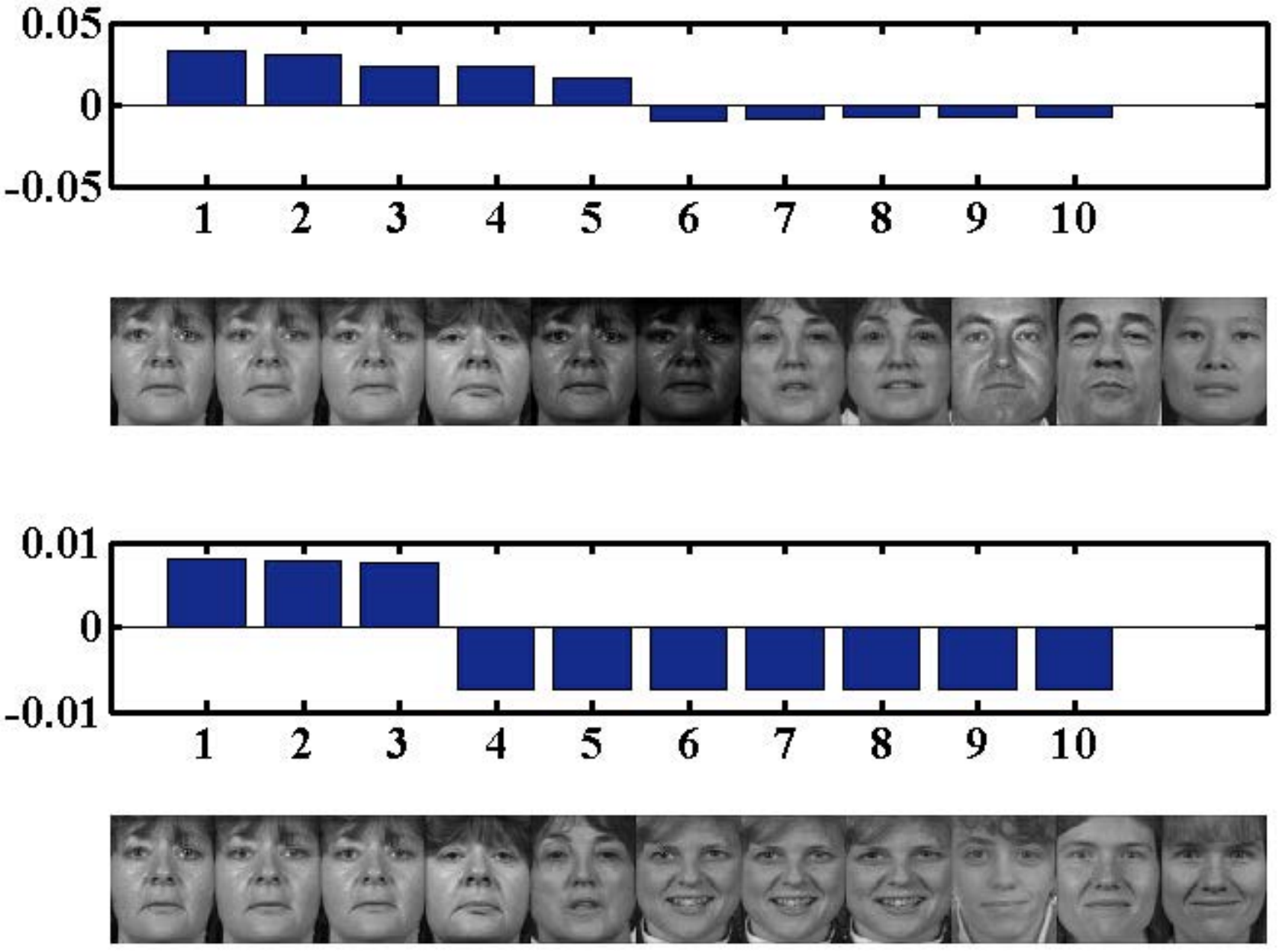}}
\subfigure[Samples with noise and their rank scores]{
\centering
\includegraphics[scale=0.45]{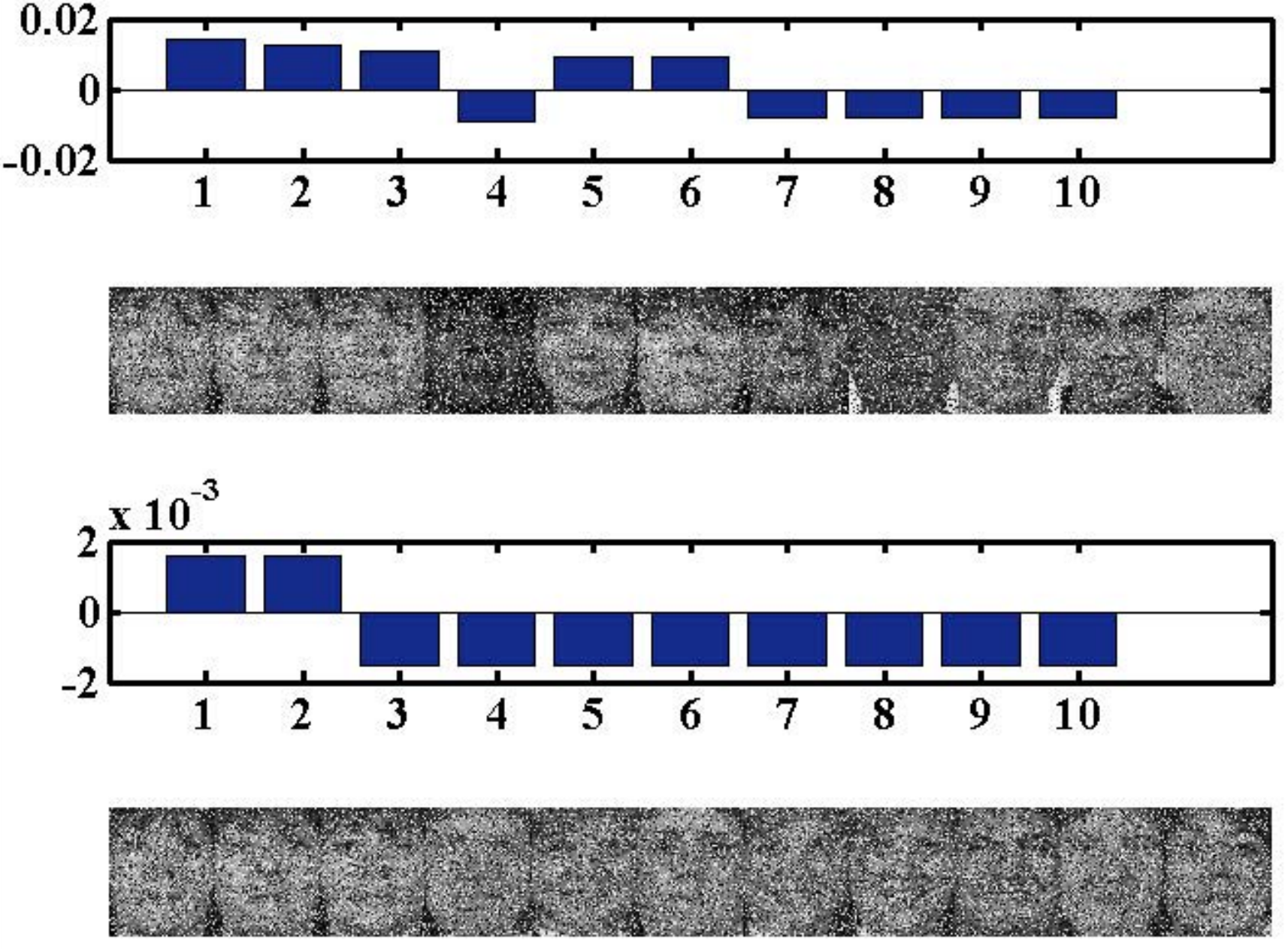}}
\caption{The figure shows the top 10 most relevant face images selected by SR and K-Nearest Neighbour based on a given query face image. This experiment is conducted in a subset of FERET database~\cite{feret} (72 subjects with 6 images in each subject). \textbf{The first two rows of the figure are the selection results of SR while the last two rows are the selection results of KNN.} The left subfigure reports the results on the original FERET database while the right one reports the results on the modified FERET database in which 30\% of pixels of each image has been corrupted by noise. In the figure, the first face image of each image array is the query image and the rest ten images are the relevant face images selected by SR or KNN. The histograms above the image array demonstrates the confidence scores of these top ten relevant face images. If the subjects of the return face image and the query face image are identical,  its corresponding histogram is positive otherwise it is negative.  In the figure, SR gets five hits either on the original FERET database or on the noisy FERET database while KNN only gets three and two hits on these two datasets respectively. Clearly, this phenomenon verifies that sparse graph, which is generated by SR, is more discriminative and robust.  }
\label{Examples}
\end{figure*}

Recently, many works leverage SR to construct a sparse graph (or $\ell_1$-graph) for tackling subspace learning, clustering and semi-supervised learning tasks~\cite{ssc,srdp,spp,sparsegraph,nlrsg}. These approaches can achieve such remarkable successes, since the sparse graph incorporates the merits of SR that it is more discriminative and robust than the conventional graph. Although a lot of impressive related works have been proposed, as far as we know, there is no prior work that directly employs the sparse graph for transduction. In this paper, we utilize the sparse graph to present a novel Graph-based Transduction (GT) algorithm for classification. Following the same graph construction manner in~\cite{srdp,sparsegraph}, each sample is taken out as the query sample and the remainder of samples are considered as the dictionary to present a Sparse Representation (SR) system in which the correlations (or similarities) between the query sample and other samples are measured. In such case, a sparse graph, which encodes the correlations between each two samples, can be constructed and it is not hard to derive its graph Laplacian. Note, the graph Laplacian is constructed from both training samples and testing samples. Finally we can achieve our proposed classification approach via plugging such graph Laplacian into the conventional Graph-based transduction framework. We name this novel graph-based classification approach Sparse Graph-based Classifier (SGC). SGC inherits the advantages of both SR and GT which is exactly the positive answer of the aforementioned question. Compared with SR, since the graph Laplacian is constructed from both training and testing samples, SGC can not only use the correlations between the given testing sample and training samples, which is as same as what the traditional SR-based classifier does, but also use the correlations of the testing data and the correlations of the training data to further improve the discriminating power of SR.  Moreover, since the testing samples are complemented to construct a larger dictionary, SGC alleviates the undercomplete dictionary issue suffered by SR~\cite{src}. Compared with GT, the graph laplacian of SGC is generated from SR instead of k-nearest-neighbour or $\epsilon$-ball. So there are two merits inherited from SR: the relevant samples can be better and adaptively selected for each sample to constitute the local clique (or neighbour); the obtained graph is more robust to noise~\cite{src,sparsegraph} (see the examples on Figure~\ref{Examples}). We apply our work to image classification. Yale~\cite{yale}, AR~\cite{ar}, FERET~\cite{feret} and Caltech256~\cite{caltech256} databases are employed for evaluation. The experimental results show that our method can get a promising result in comparison with the state-of-the-art classifiers particularly in the small training sample size case.

The rest of paper is organized as follows: Section~\ref{s2} presents the related works; Section~\ref{s3} describes the proposed approach. Section \ref{s4} shows the experimental evaluation of our works; the conclusion is summarized in Section~\ref{s5}.

\section{Related Works}
\label{s2}
\subsection{Sparse Representation}
Sparse Representation (SR) is a hot topic in the recent decade and widely applied to extensive areas~\cite{src,ksrc,sparsegraph,srod,mjsr,srdp}. Since SR enjoys the good discriminating power and the robustness to noise, SR is often considered as a popular classification technique. For example, Wright et al considered the testing face image as a query and the training face images as the visual dictionary to construct a SR system to address the face recognition task~\cite{src}. Gao et al kernelized the previous approach and apply the kernel version to the face recognition and image classification~\cite{ksrc}. To overcome the undercomplete dictionary situation and further improve the performance of SR-based face recognition, Ma et al~\cite{gsrc} complemented the visual dictionary by adding the gradient image of the faces. However, the original faces and gradient faces are totally in the different feature domains. Agarwal e al introduced a work for learning a sparse, part-based representation for object detection~\cite{srod}.  Yuan et al presented a multitask joint sparse representation model to combine the strength of multiple features and/or instances for visual classification~\cite{mjsr}. Although these SR-based approaches achieve remarkable successes, there are two main shortcomings which are still not essentially overcame. The first one is that SR cannot perform well in the undercomplete dictionary case (the small training sample size case). The second shortcoming is that conventional SR only can utilize the correlations (or similarities) between training samples and the testing samples to infer the class label while cannot sufficiently exploit the correlations of the training samples as well as the correlations of the testing samples. The proposed Sparse Graph-based Classifier (SGC) can well overcome these two shortcomings.
\subsection{Sparse Graph}
\label{sg}
Motivated by the recent successes of SR~\cite{src,ksrc,wsrc}, some researchers leverage SR instead of the conventional k-nearest-neighbour or $\epsilon$-ball to construct a sparse graph for addressing the different issues~\cite{ssc,srdp,sparsegraph,spp,nlrsg}. More specifically, Qiao et al and Timofte et al successively use SR to construct a sparse graph for dimensionality reduction~\cite{srdp,spp}. The Sparse Subspace Clustering (SSC) algorithms~\cite{ssc,sssc,rss} learn a sparse graph for clustering via considering the data self representation problem as a SR issue. Similar to~\cite{srdp,spp}, Cheng et al utilize SR to construct the $\ell_1$-graph (sparse-graph) for spectral clustering, subspace learning and semi-supervised learning~\cite{sparsegraph}. Although the applications and the learning (or construction) procedures of these works are very different, the obtained sparse graphs are very similar which all demonstrate the good discriminative abilities and robustness. In this paper, we intend to use the sparse graph to present a GT algorithm which can incorporate these desirable properties. As same as these works~\cite{srdp,sparsegraph,spp,nlrsg}, our approach is also an application of the sparse graph.
\subsection{Graph-based Transduction}
As a transductive learning algorithm, Graph-based Transduction (GT) labels the samples based on the similarities between each two sample (no matter the training sample or the testing samples) where these similarities are encoded in a graph (or hypergraph). In other words, GT can sufficiently exploit the information of whole data and therefore it often performs well in the small training sample size case. This fact makes GT become very popular approach for classification and labeling~\cite{hyper,zhou,st,gtam,gtc,srgt}. For example, Duchenne et al presented a state-of-the-art segmentation via leveraging the conventional GT to infer the label of each pixel~\cite{st}. Graph Transduction via Alternating Minimization (GTAM) enhanced GT via introducing a propagation algorithm, which can more reliably minimize a cost function over both a function on the graph and a binary label matrix, and applying it to classification~\cite{gtam}. Similarly, in order to address the classification issue, Orach et al presented a new GT algorithm via introducing an additional quantity of confidence in label assignments and learning them jointly with the weights~\cite{gtc}. Zhou et al provided a new way to construct the hypergraph and used it to replace the graph in the GT framework for tackling a labeling task~\cite{zhou}. Following the same framework in~\cite{zhou}, Yu et al presented a GT-based image classification approach via adaptively generating the hyperedges and learning their weights~\cite{hyper}. From this short review, it is not hard to conclude that one of the important factors to effect the success of GT algorithm is the quality of the graph (or hypergraph). The graphs (or hypegraphs) of aforementioned approaches are generated by k-nearest-neighbour or $\epsilon$-ball. However, some works have indicated that such graphs can be easily corrupted by noise~\cite{sparsegraph} (see the examples in Figure~\ref{Examples}). Inspired by the approaches mentioned in Section~\ref{sg}~\cite{srdp,spp,sparsegraph}, in our approach, we adopt the more robust and discriminative graph, sparse graph, to alleviate this problem.

\section{Methodology}
\label{s3}
\subsection{Sparse Graph Laplacian}
\label{cgraphp}
The graph plays a very important role in Graph-based Transduction (GT), since it depicts the \emph{relationships} (similarities or correlations) of the samples which are regarded as the basis for classification (or labeling). However, the conventional GT approaches generate the graphs (or hypergraph) by k-nearest-neighbour or $\epsilon$-ball. It has been proved that these graphs often cannot well reveal the real relationships of samples due to noise and some other factors~\cite{sparsegraph}. Some recent works~\cite{srdp,sparsegraph,spp} indicate that using the Sparse Representation (SR) can generate a more discriminative and robust graph. So, in this section, we will introduce how to use SR to construct a high quality graph. Following the same graph construction manner in~\cite{srdp,sparsegraph}, we take out one sample from the whole dataset and consider the rest samples as the dictionary to construct a SR system. Here, we let $d\times n$-dimensional matrix, $X=[x_1,\cdots,x_i,\cdots,x_n]$, be the sample matrix where $d$ is the dimension of sample and $n$ is the number of samples. We denote the sample that we want to represent, $x_q$, where $q$ is its corresponding index. The matrix $X_{i\neq q}=[x_1,\cdots,x_{q-1},x_{q+1},\cdots,x_n]$ is the sample matrix which excludes the sample $x_q$. The correlations (or similarities) between the query sample $x_q$ and the other samples are measured by solving the following SR problem
\begin{equation}\label{l0issue}
  \hat{c}_q=\arg\underset{c_q}\min||c_q||_0,~\text{s.t.}~ ||x_q-X_{i\neq q} c_q^T||^2\leq \epsilon
\end{equation}
where the vector $c_q=[c_{q}(1),\cdots,c_{q}(i-1),c_{q}(i+1),\cdots,c_{q}(n)]^T$ is the representation coefficients (regression weights) of sample $x_q$ and $c_{q}(t)$ is the element of $c_q$ corresponding to the sample $x_t$. $\epsilon$ is the measurement noise. However, this $\ell_0$-norm constrained representation issue is NP-hard and difficult even to approximate~\cite{src,nphard}. Only a few of very recent works attempt to solve the problem as a non-convex minimization issue~\cite{tpm,l0debur}, and some of these works even cannot guarantee the converge. The researchers more tend to seek the close-form solution via considering this $\ell_0$-norm constrained regression problem as a $\ell_1$-norm constrained problem
\begin{eqnarray}\label{l1norm}
  &&\hat{c}_q=\arg\underset{c_q}\min||c_q||_1,~\text{s.t.}~ ||x_q-X_{i\neq q} c_q^T||^2\leq \epsilon\\ \nonumber
  &\Rightarrow&\hat{c}_q=\arg\underset{c_q}\min\{(1-\beta)||x_q-X_{i\neq q} c_q^T||^2+\beta||c_q||_1\}
\end{eqnarray}
where $\beta$ is a parameter in the range $[0, 1]$ which is used to control the trade off between the reconstruction error and the sparsity. This problem is a typical convex problem. So it can be solved by many mature convex optimization techniques. Moreover, another reason that $\ell_1$-norm may be more suitable to construct a high quality sparse graph is that, unlike the $\ell_0$-norm, which only counts the nonzero elements of coefficients, $\ell_1$ also pays attention on the values of coefficients which indicate the degrees of similarities. Of course, the idea of the sparse graph is general. So other norms can also be applied to construct some other graphs which incorporate different specific properties.

In our model, we adopt the SLEP method~\cite{slep} to efficiently solve the problem in Equation~\ref{l1norm}.
The correlation between sample $x_i$ and $x_j$, which is also regarded as the weight of edge between $x_i$ and $x_j$, can be calculated as follows
 \begin{equation}\label{}
  w_{ij}=w_{ji}=\frac{|c_{i}(j)|+|c_{j}(i)|}{2}
\end{equation}
where $w_{ij}$ is also the ($i,j$)-th element of affinity matrix of sparse graph, $W$. Moreover, we define the self-similarity of the sample as follows
 \begin{equation}\label{}
  w_{ii}=\sum_{t\neq i}w_{it}
\end{equation}
We use the Laplacian Eigenmapping~\cite{laplacian} to derive the graph Laplacian. The normalized graph Laplacian can be computed as follows
\begin{equation}\label{}
L=D^{-1/2}(D-W)D^{-1/2}=I-D^{-1/2}WD^{-1/2}
\end{equation}
where $D$ is a diagonal matrix and $D_{ii}= \sum_{j}w_{ij}$. $I$ is an identical matrix. This normalized graph Laplacian incorporates the properties of SR which is more discriminative and enjoys the robustness to noise.

\subsection{Sparse Graph-based Transduction}
Graph-based transduction (GT) methods label input data by learning a classification function that is regularized to exhibit smoothness along a graph over labeled and unlabeled samples~\cite{zhou,gtam}. In other words, the GT model can be deemed as a regularized graph cut problem in which the graph cut is considered as a classification function. Based on the obtained sparse graph Laplacian $L$, we first formulate our GT method in the binary class case and then generalize it into the multi-class case. Since our method is based on sparse graph, we name our proposed GT algorithm Sparse Graph-based Transduction (SGT) and its corresponding classifier Sparse Graph-based Classifier (SGC). In SGC, a graph cut $f$ is defined as the classification function and this cut should not only minimize the similarities losses (sparse representation relationship losses) but also reduce the classification errors of the training samples. Mathematically speaking, such model can be formatted as follows
\begin{eqnarray}\label{binary}
\hat{f}&=&\arg\underset{f}\min\{ \Omega(L,f)+\lambda \Phi(y,f)\}\\ \nonumber
&=&\arg\underset{f}\min\{f^TLf+\lambda||f-y||^2 \}
\end{eqnarray}
where the similarity loss function $\Omega(L,f)=f^TLf$ is denoted as a normalized cut function~\cite{ncut} and the classification error function $\Phi(y,f)=||f-y||^2$ measures the classification errors by computing the Euclidean distances between the predicted labels and groundtruth labels. The vector $y$ is the label vector. Let us assume $y(i)$ is the $i$-the element of $y$, which depicts the status of the sample $x_i$. Then, in $y$, $y(i) = 1$ or -1 if the sample $x_i$ has been labeled as positive or negative respectively, and 0 if it is unlabeled. $\lambda$ is a positive to reconcile the similarity losses, $\Omega(L,f)$, and the classification errors, $\Phi(y,f)$. Note, the graph Laplacian $L$ is constructed from both training samples and testing samples.  Moreover, it is worthwhile to point out that the GT framework is very flexible. The researchers can also design these two loss functions by themselves for addressing different issues.

We employ the one-versus-all strategy to generalize the algorithm from the binary classification case to the multi-class classification case. The multi-class version is denoted as follows
\begin{eqnarray}\label{multi}\nonumber
  &&\hat{F}=\arg\underset{\hat{F}}\min \sum_i \{ \Omega(L,f_i)+\lambda \Phi(y_i,f_i)\}\\
&\Rightarrow&\hat{F}=\arg\underset{\hat{F}}\min \sum_i \{ f_i^T L f_i +\lambda ||f_i-y_i||^2 \}\\ \nonumber
&\Rightarrow&\hat{F}=\arg\underset{F}\min \{ F^TLF+\lambda||F-Y||^2\}
\end{eqnarray}
where $F=[f_1,\cdots,f_i,\cdots,f_c]$ and $Y=[y_1,\cdots,y_i,\cdots,y_c]$ are the collection of classification functions and the collection of the defined labels with respect to the different classes. $c$ is the number of classes. In label vector $y_i$, only the samples from class $i$ are considered as positive while the samples from other classes are considered as negative.

Since $L$ is a positive semi-definite matrix, Equation~\ref{multi} can be efficiently solved by Regularized Least Square (RLS). We obtain the partial derivative of Equation~\ref{multi} with respect to $F$, and let it equal to zero.
\begin{eqnarray}\label{}\nonumber
&~~~~&\frac{\partial}{\partial F} \left\{ F^TLF+\lambda||F-Y||^2 \right\}=0\\
&\Rightarrow& 2(LF+\lambda F-\lambda Y)=0\\\nonumber
&\Rightarrow& F=\frac{\lambda Y}{L+\lambda I}
\end{eqnarray}
Finally, the classification of $i$-th sample can be accomplished by assigning it to the $j$-th class that satisfies
\begin{equation}\label{}
\hat{j}=\arg\underset{j}\max F_{ij}.
\end{equation}
where $F_{ij}$ is the ($i,j$)-th element of matrix $F$.

SGT inherits the desirable properties of both GT and SR. More specifically, SGC can well exploit the correlations of both testing samples and the training samples, since the graph Laplacian is constructed from whole data. SGC performs much better in the small training sample size case, since SGC utilizes the testing samples to complement the dictionary of SR in the sparse graph construction procedure. Moreover, SGC is more discriminative and robust to noise.
\section{Experimental Results}
\label{s4}
\begin{figure}[!tbp]
\centering
\subfigure[Yale]{
\centering
\includegraphics[scale=0.28]{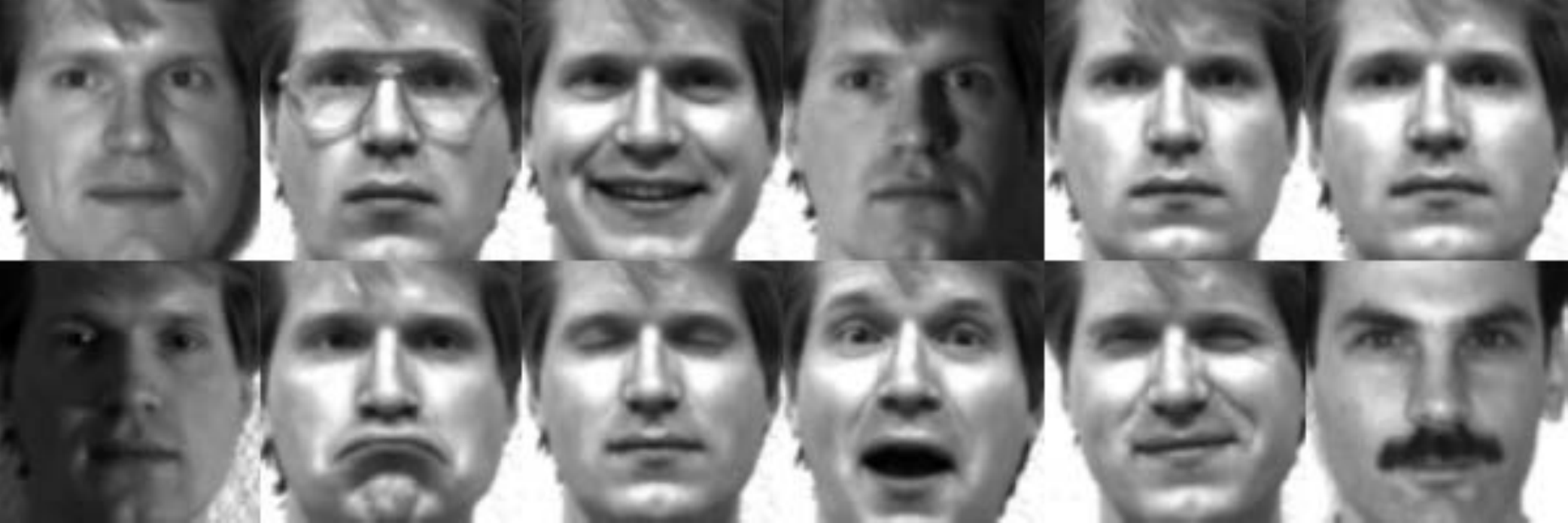}
}
\subfigure[AR]{
\centering
\includegraphics[scale=0.55]{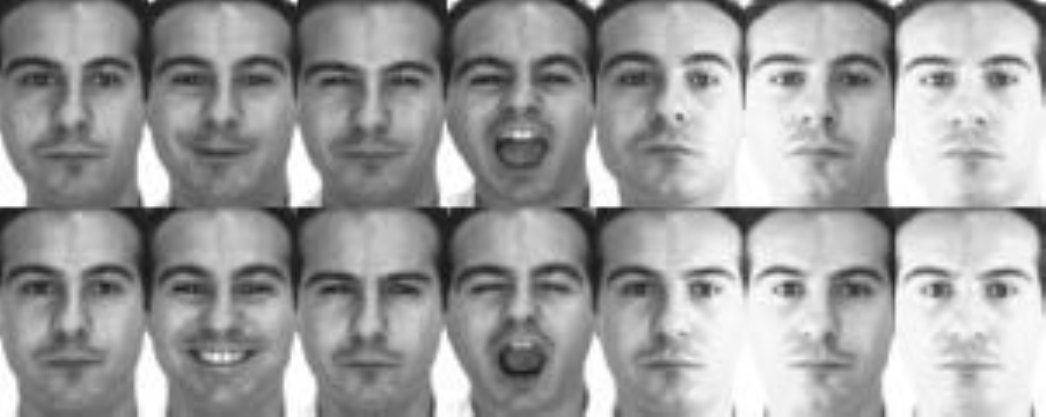}
}
\subfigure[FERET]{
\centering
\includegraphics[scale=0.34]{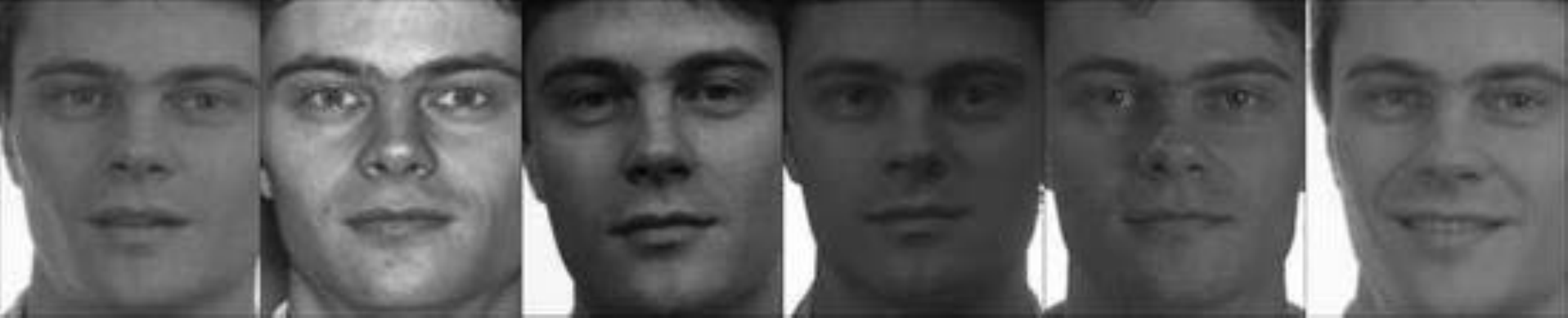}
}
\subfigure[Caltech256]{
\centering
\includegraphics[scale=0.32]{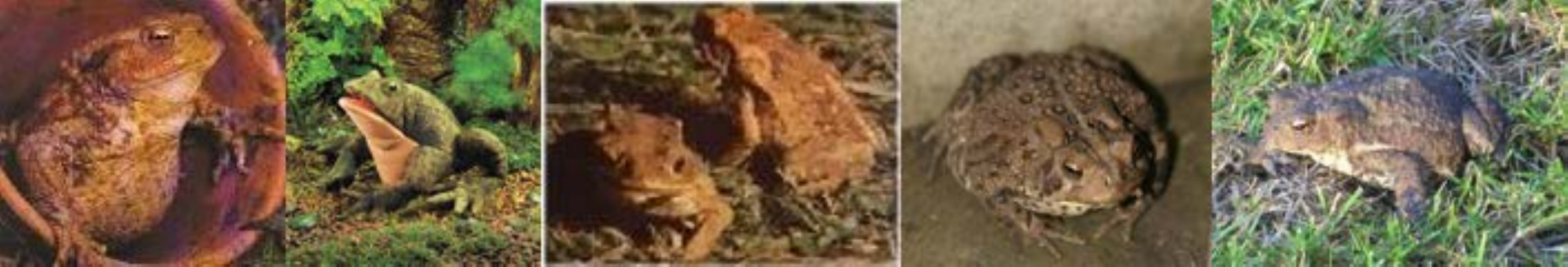}
}
\caption{The sample images of the datasets used in our experiments.}
\label{dataset}
\end{figure}

Yale~\cite{yale}, FERET~\cite{feret}, AR~\cite{ar} and Caltech256~\cite{caltech256} databases are used to evaluate our work. The Yale face database totally has 15 subjects and 11 samples per subject~\cite{yale}. The size of image is 32$\times$32 pixels. The FERET database contains 13539 images corresponding to 1565
subjects~\cite{feret}. Following paper~\cite{dhlp}, a subset which contains 436 images of 72 individuals is selected in our experiments and this subset involves variations of facial expression, illumination and poses. The AR database consists of more than 4,000 images of 126 subjects \cite{ar}. The database characterizes divergence from ideal conditions by incorporating various facial expressions, luminance alterations, and occlusion modes. Following paper \cite{lrc}, a subset contains 1680 images with 120 subjects are constructed in our experiment. All these images are 50$\times$40 pixels. Similarly, we follow the paper~\cite{hyper} and select a subset from Caltech256 database~\cite{caltech256}. In this subset, there are 20 classes and 100 images per class. Since Caltech256 is more challenging than the other two databases. We adopt the Picodes feature~\cite{picodes} to represent the images. The AR, Yale and FERET databases are the face databases and Caltech256 is the image databases. Figure~\ref{dataset} shows some example images of these databases.

Sparse Representation-based Classifier (SRC)~\cite{src}, Collaborative Representation-based Classifier (CRC)~\cite{crc}, LIBSVM~\cite{libsvm}, Graph-based Classifier (GC) (The corresponding Classifier of Graph-based Transduction (GT) algorithm~\cite{st,gtam}), Normalized Hypergraph-based Classifier (NHC)~\cite{zhou} and Adaptive Hypergraph-based Classifier (AHC)~\cite{hyper} are employed for comparison. The last three algorithms are all transductive learning-based methods and their graph matrices (or hypergraph matrices) are generated based on Euclidean distance (Heat Kernel Weighting).

\begin{table*}[!tbp]
\caption{Two-fold cross validation results on Yale, AR and Caltech256 databases. \label{results}}
\footnotesize
\begin{center}
    \begin{tabular}{p{1.4cm}<{\centering}|cccccccc}
    \hline
     \multirow{2}*{Databases}
    &\multicolumn{7}{c}{Classification Error (Mean$\pm$STD,\%)}\\ \cline{2-8}
    &CRC~\cite{crc}&LIBSVM~\cite{libsvm}&SRC~\cite{src}&NHC~\cite{zhou}&AHC~\cite{hyper}&GC&\textbf{Ours} \\
    \hline
Yale&9.33$\pm$5.66&10.67$\pm$3.77&3.33$\pm$2.83&20.00$\pm$7.86&20.56$\pm$7.07&8.89$\pm$6.29&\textbf{2.78$\pm$2.36}\\
AR&31.25$\pm$0.42&31.90$\pm$0.67&27.98$\pm$2.36&31.61$\pm$0.76&31.73$\pm$0.08&42.08$\pm$2.44&\textbf{26.13$\pm$0.42}\\
Caltech256&\textbf{38.20$\pm$1.55}&39.15$\pm$2.47&40.85$\pm$2.19&43.00$\pm$2.40&43.10$\pm$2.26&52.50$\pm$4.10&39.40$\pm$3.45\\
FERET&15.28$\pm$0.00&12.96$\pm$0.56 &11.34$\pm$0.03&35.42$\pm$1.64&33.10$\pm$2.29&11.11$\pm$1.31&\textbf{10.19$\pm$0.65} \\
\hline
Average&23.52&23.67&20.88&32.51&32.12&28.65&\textbf{19.63}\\
      \hline
    \end{tabular}{}
    \end{center}
\end{table*}

\subsection{Image Classification}
We apply different classifiers to these four databases and the two-fold cross validation is adopted in our experiments. Table~\ref{results} reports the classification results. From the observations, we can know that the proposed Sparse Graph-based Classifier (SGC) outperforms all the compared classifiers on AR, Yale and FERET databases and can get a very promising performance on Caltech256 database. Moreover, SGC improves the performances of both SRC and GT-based algorithms (NHC, AHC and GC). For example, the classification accuracy gains of SGC over SRC, NHC, AHC and GC are 1.25\%, 12.88\%, 12.49\% and 9.02\% respectively in average. In the experiments, the GT-based algorithms perform not well in comparison with SRC and SGC. We think there are two reasons behind this phenomenon. The first reason is that k-nearest-neighbour is not discriminative enough to well select the relevant samples. The second reasons is that it is hard to select a suitable $k$ to define a suitable neighbour, which can well reveal the local relationships of samples, while SGC can avoid such selection of $k$, since the relevant samples are adaptively selected (without giving any $k$). We observe from the classification performances of SRC and SGC that the performance of SGC relies on the performance of SRC. This phenomenon verifies that the core of SGC is the sparse graph which is generated by SR and incorporates the properties of SR.

\subsection{Robustness to Noise}
\begin{figure}[h]
\centering
\subfigure[FERET]{
\centering
\includegraphics[scale=0.35]{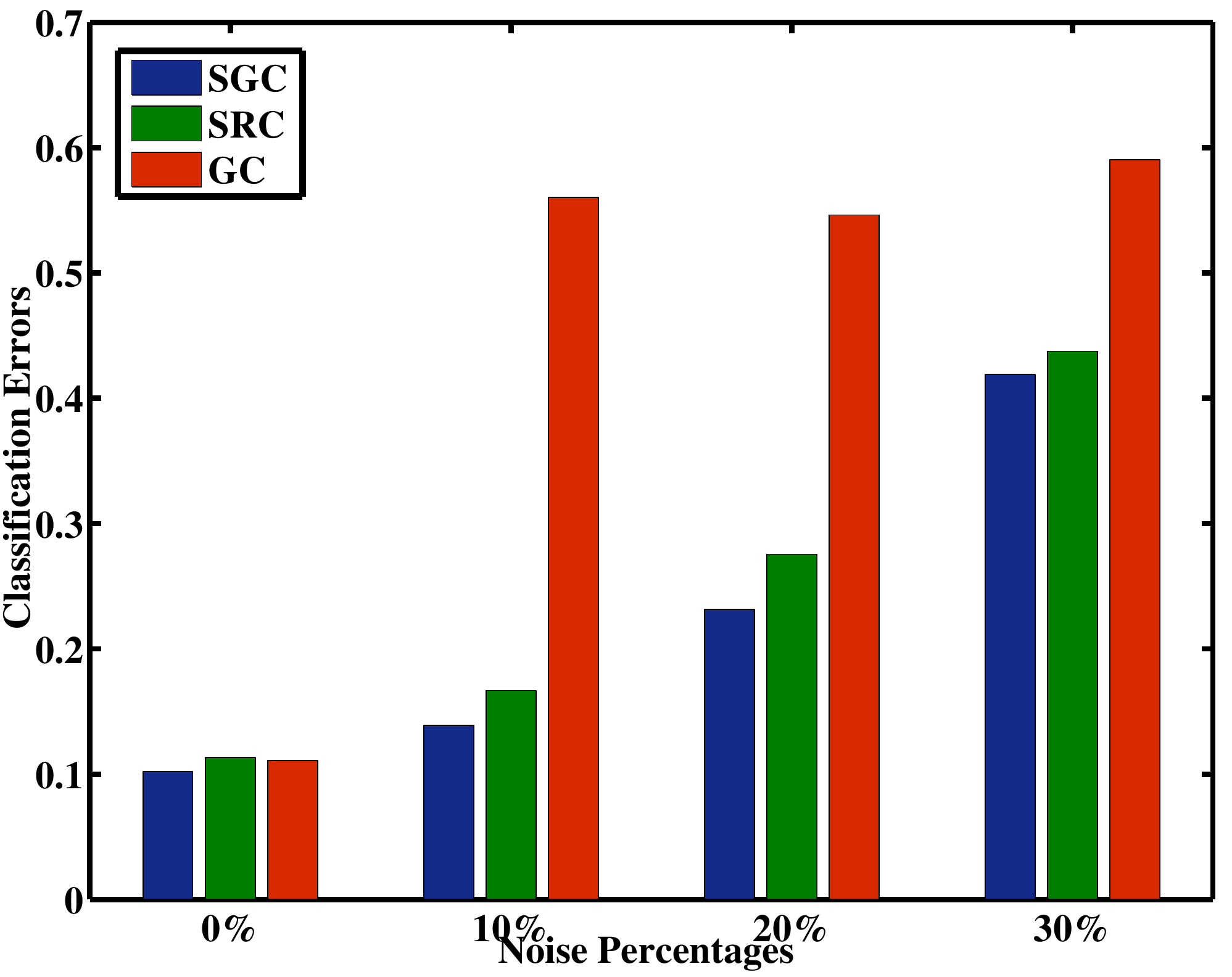}
}
\subfigure[AR]{
\centering
\includegraphics[scale=0.35]{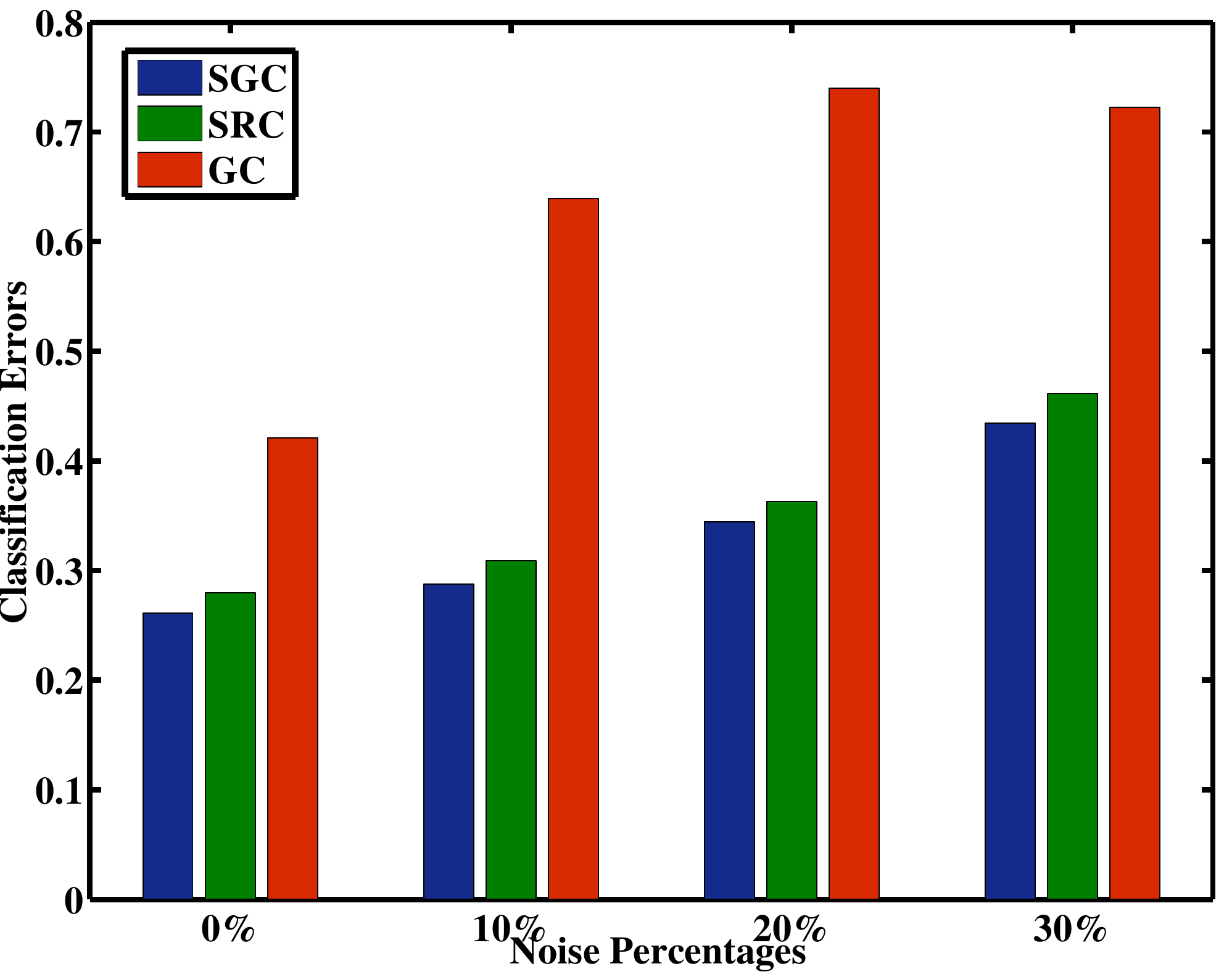}
}
\caption{Classification performances of different approaches under different noise levels.}
\label{noiselevel}
\end{figure}

In SGC, the graph Laplacian is generated by SR. So SGC should inherit some merits from SRC. Theoretically speaking, compared to the original GT approaches, SGT should be more robust to noise. In this section, we conduct some experiments on AR and FERET databases to validate this. In the experiments, several noisy face databases are constructed by randomly generating the salt-and-pepper noise for each face image. We define four noise levels based on the proportion of noise in a image and study the effect of noise proportion to the classification performance. As same as the experimental setting in the previous section, the two-fold cross-validation is adopted to measure the classification performance. Figure~\ref{noiselevel} shows the experimental results. In this figure, the $x$-axis indicates the proportion of noise and the $y$-axis indicates the misclassification accuracy. From the figure, SGC outperforms SRC and GC in all experiments and has a similar behaviour as SRC. GC fails soon even in the case that only 10\% noise is introduced. On the contrary, the classification performances of SRC and SGC drop slowly along with the increasing of the noise percentage. Clearly, such phenomenon well verifies that SGC is more robust to noise in comparison with the conventional GT algorithms.

\subsection{Insensitivity to the Training Sample Size }
\begin{figure}[h]
\centering
\subfigure[Yale]{
\centering
\includegraphics[scale=0.35]{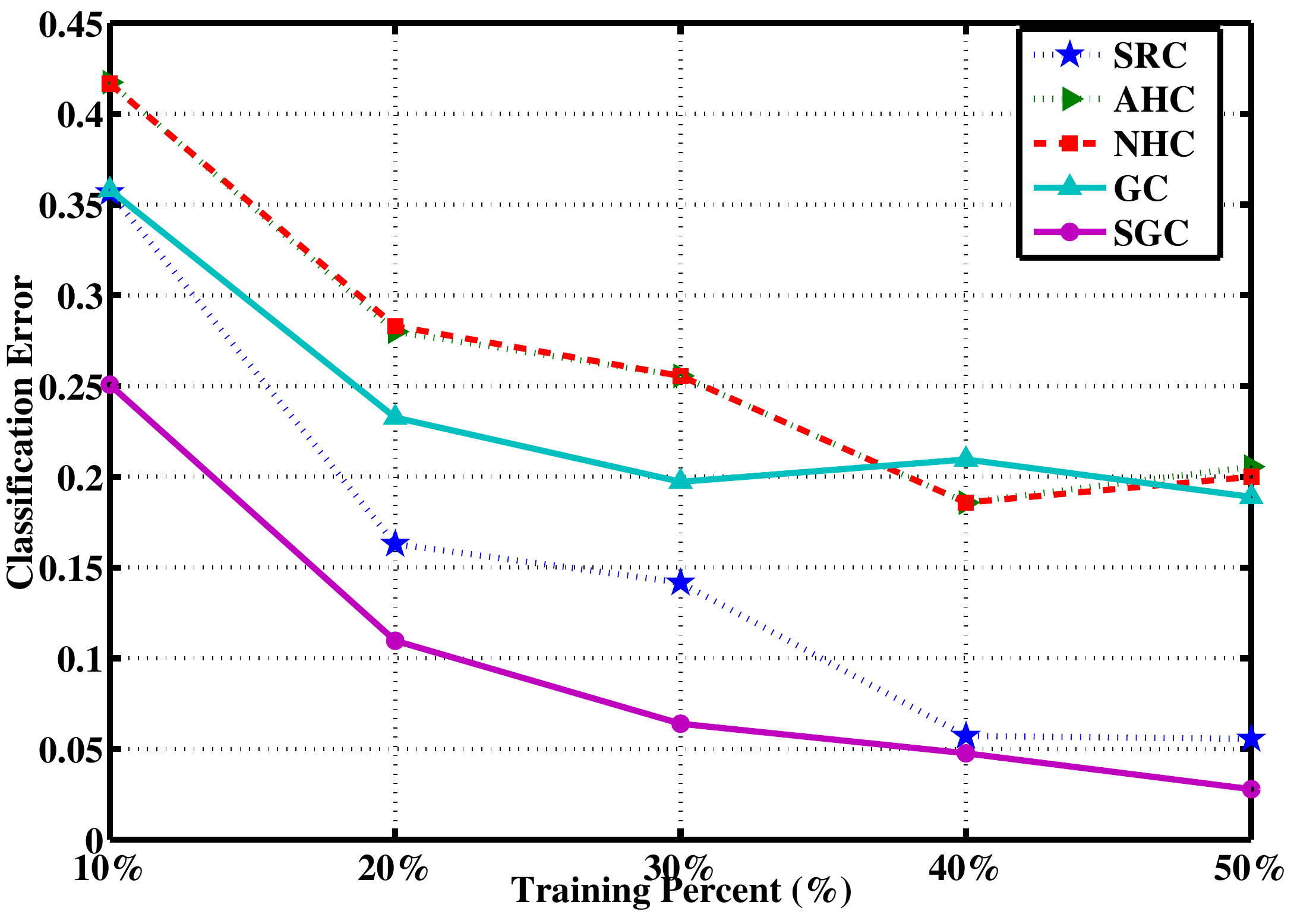}
}
\subfigure[AR]{
\centering
\includegraphics[scale=0.35]{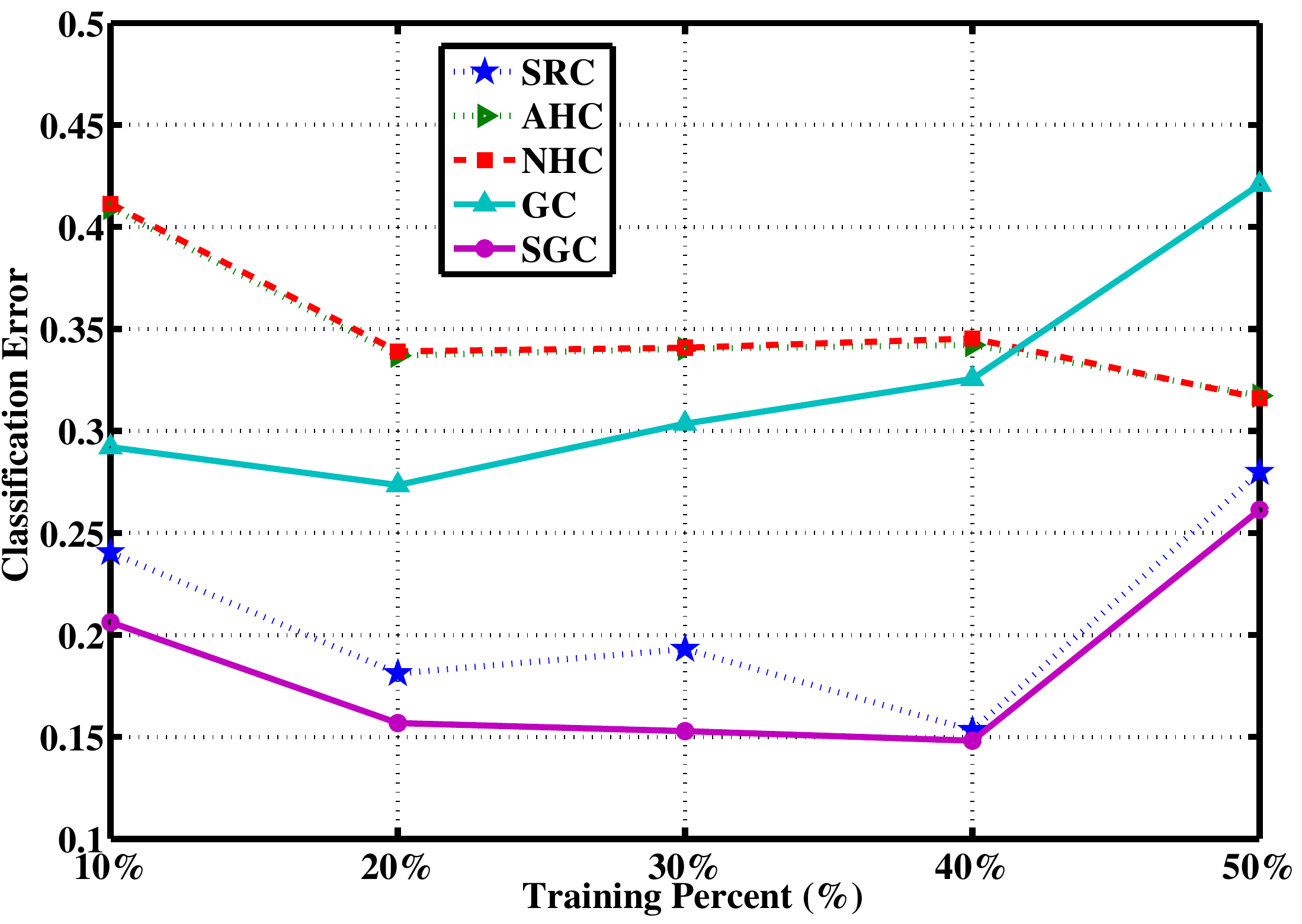}
}
\caption{Classification errors of different methods under different training sample sizes.}
\label{trainratio}
\end{figure}

The main advantage of GT approaches is that the information of both training data and testing data can be fully exploited. So, in most of time, these approaches always perform much better than other approaches in small training sample size case. As an instance of the GT framework, SGT should also have such desirable property. In this section, we conduct several experiments on AR and Yale databases to investigate the effect of the training sample size to the classification performance of SGC. In these experiments, the cross-validation strategy is employed for measuring the classification performance and five sizes of training samples are defined. For example, if the proportion of the training sample is 0.1, we adopt ten-fold cross-validation to conduct the experiments. We plot the classification errors of different approaches under different training sample sizes in Figure~\ref{trainratio}. The $x$-axis indicates the training sample percentage of data and the $y$-axis indicates the mean classification error. From the observations in Figure~\ref{trainratio}, SGC consistently outperforms SRC and the improvement of SGC over SRC is increased along with the reduction of the training proportion. These phenomena all verify that SGC can perform much better than SRC in the small training sample size case.

\subsection{The Settings of Parameters}
\begin{figure}[!tbp]
\centering
\subfigure[The effect of $\beta$]{
\centering
\includegraphics[scale=0.38]{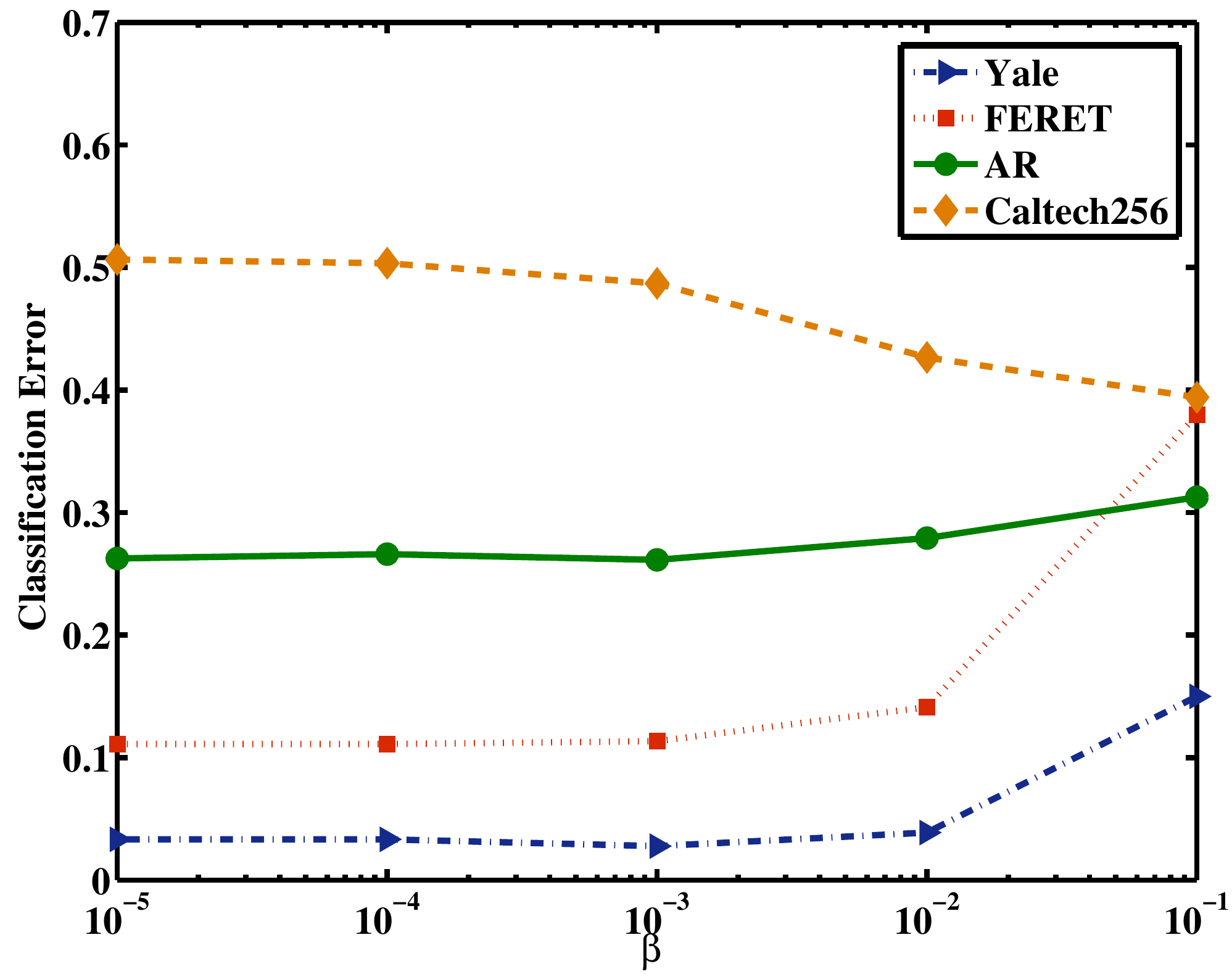}
}
\subfigure[The effect of $\lambda$]{
\centering
\includegraphics[scale=0.38]{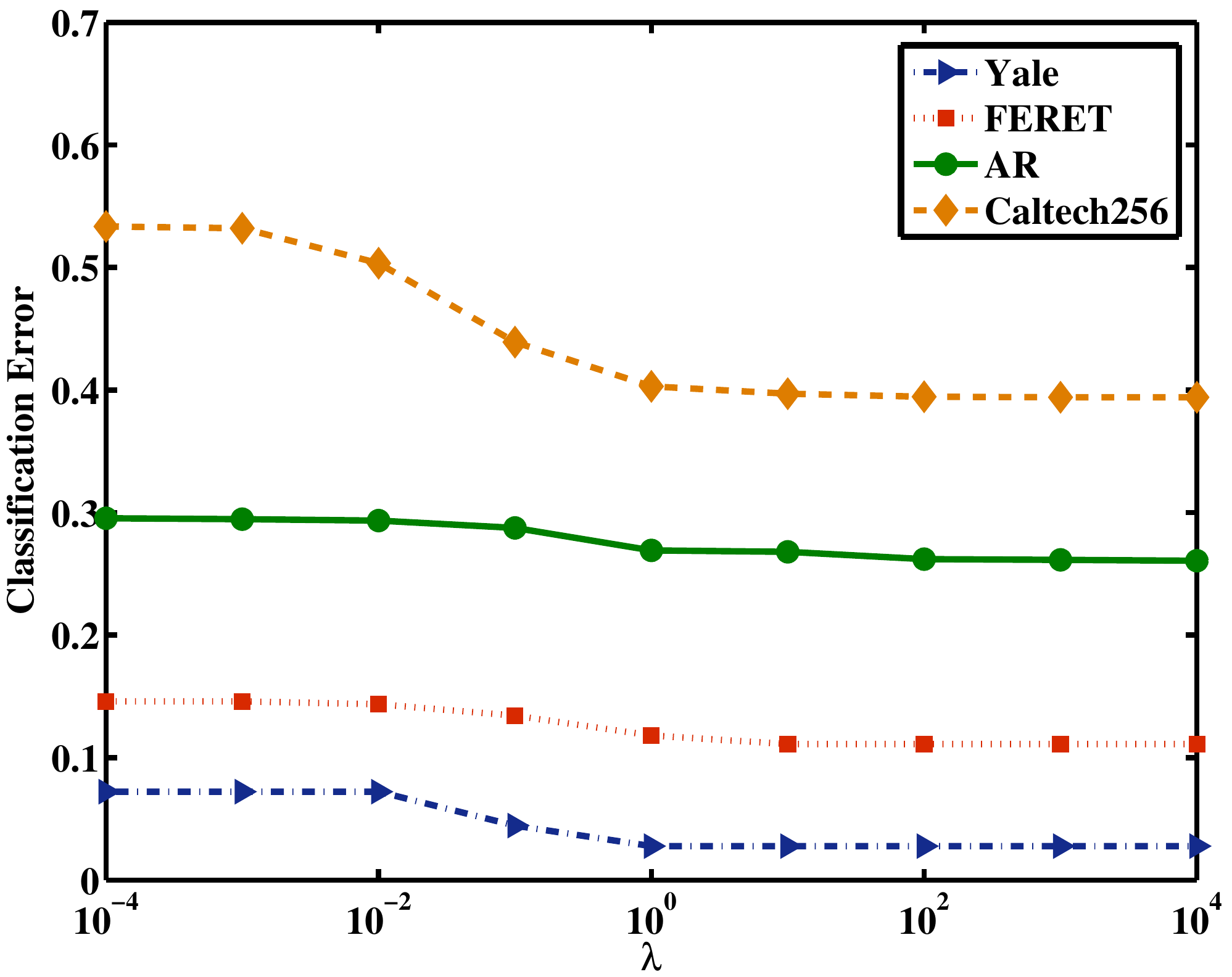}
}
\caption{The effects of parameters to the classification performance.}
\label{Params}
\end{figure}

There are two parameters in SGC. One is $\beta$ which is introduced by SR and used to control the degree of sparsity. The other is $\lambda$ which is used to reconcile the correlation loss and classification errors of training samples. In this section, we conduct some experiments to discuss the effects of these parameters to the classification performance. As same as the previous section, the two-fold cross-validation is adopted. Figure~\ref{Params} plots the relationships between the classification error and the value of parameters.  From this figure, we can find that SGC is quite insensitive to $\beta$ on three face databases, namely Yale, AR and FERET, when $\beta\leq 10^{-2}$. So we suggest that the optimal $\beta$ of these face databases are all equal to $10^{-3}$. However, for Caltech256 database, the optimal $\beta$ is much greater and its value is $0.1$. The settings of $\beta$ on Caltech256 database is different to the ones on three face databases, since their features are different. The feature of the face databases is just the simple gray scale while the feature of the Caltech256 database is Picodes. Similarly, SGC is quite insensitive to $\lambda$ when its value is greater than 1. From the observations, we can conclude that the optimal $\lambda$ for all four databases is $10^{3}$.

\section{Conclusion}
\label{s5}
We introduced the Sparse Representation (SR) to the Graph-based Transduction (GT) and presented a novel GT-based Classifier called Sparse Graph-based Classifier (SGC) for image classification. In SGC, SR is utilized to measure the correlation of each two samples. Then a sparse graph is constructed to depict such correlations. Finally, the graph Laplacian of this graph is plugged into the GT framework to infer the labels of the unlabeled samples. According to the theoretical analysis and the experimental verification on four popular image databases, we concluded that SGC can incorporate the advantages of both SR and GT. SGC is a very flexible framework, since its parts are all replaceable. So there are a lot of interesting works that can be done based on SGC. For example, if we want to enhance SGC, we can design the classification error function $\Phi(y,f)$ by ourselves or utilize other more advanced regression techniques to instead of SR to construct the high quality graph.

\section*{Acknowledgement}
The work described in this paper was partially supported by National Natural Science Foundations of China (NO. 60975015 and 61173131), Fundamental Research Funds for the Central Universities (No. CDJXS11181162). The authors would like to thank useful comments of the anonymous reviewers and editors.


\bibliographystyle{elsarticle-num}
\bibliography{mybib}


%


\end{document}